%% file: main.tex
\documentclass[10pt,twocolumn,letterpaper]{article}

\usepackage{wacv}

\usepackage{times}
\usepackage{epsfig}
\usepackage{graphicx}
\usepackage{amsmath}
\usepackage{amssymb}
\usepackage{subfig}
\usepackage{algorithm}
\usepackage{algorithmic}
\usepackage{booktabs} 
\usepackage{multirow}
\usepackage{adjustbox}
\usepackage{arydshln}
\usepackage{amsmath}
\usepackage{verbatim}
\usepackage[flushleft]{threeparttable}
\usepackage[table, svgnames, dvipsnames]{xcolor}

\usepackage[accsupp]{axessibility}  

\def\wacvPaperID{897} 

\wacvfinalcopy 

\ifwacvfinal
\fi

\ifwacvfinal
\usepackage[breaklinks=true,bookmarks=false]{hyperref}
\else
\usepackage[pagebackref=true,breaklinks=true,colorlinks,bookmarks=false]{hyperref}
\fi

\begin{document}

\title{Batch Normalization Tells You Which Filter is Important}

\author{Junghun Oh\\
\and
Heewon Kim\\
\and
Sungyong Baik\\
\and
Cheeun Hong
\and
Kyoung Mu Lee \vspace{0.1cm}\\
\hspace{-13cm}{Department of ECE, ASRI, Seoul National University}\\
\hspace{-13cm}{\tt\small \{dh6dh, ghimhw, dsybaik, cheeun914, kyoungmu\}@snu.ac.kr}\\
}

\maketitle

\ifwacvfinal
\thispagestyle{empty}
\fi

\input{main/abstract}

\input{main/introduction}
\input{main/related_work}
\input{main/method}

\input{main/experiment}
\input{main/conclusion}

\paragraph{Acknowledgment}    
\noindent This work was supported in part by IITP grant funded by the Korean government (MSIT) [NO.2021-0-01343, Artificial Intelligence Graduate School Program (Seoul National University)]

{\small
\bibliographystyle{ieee_fullname}
\bibliography{egbib}
}

\end{document}

%% file: main/abstract.tex
\begin{abstract}
The goal of filter pruning is to search for unimportant filters to remove in order to make convolutional neural networks (CNNs) efficient without sacrificing the performance in the process.
The challenge lies in finding information that can help determine how important or relevant each filter is with respect to the final output of neural networks.
In this work, we share our observation that the batch normalization (BN) parameters of pre-trained CNNs can be used to estimate the feature distribution of activation outputs, without processing of training data. 
Upon observation, we propose a simple yet effective filter pruning method by evaluating the importance of each filter based on the BN parameters of pre-trained CNNs.
The experimental results on CIFAR-10 and ImageNet demonstrate that the proposed method can achieve outstanding performance with and without fine-tuning in terms of the trade-off between the accuracy drop and the reduction in computational complexity and number of parameters of pruned networks.

\end{abstract}

%% file: main/introduction.tex
\section{Introduction}\label{intro}

\begin{figure}[t]
\begin{center}
\includegraphics[width=1\linewidth]{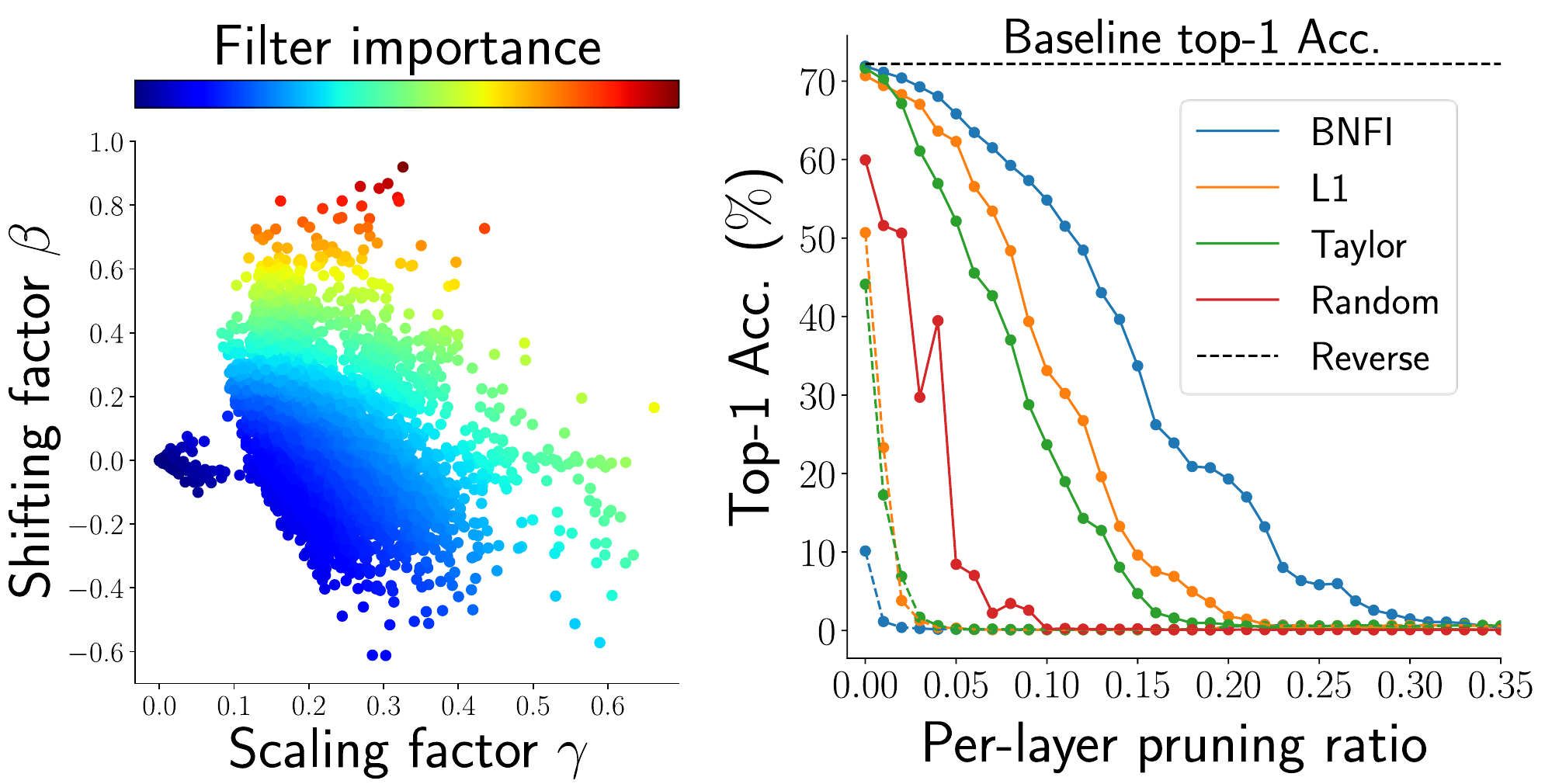}
\end{center}
\begin{center}
\end{center}
\vspace{-1cm}
\caption{
Our proposed filter importance measure is illustrated in the left figure.
The figure visualizes how pre-trained batch normalization (BN) parameters, $\gamma$ and $\beta$, are mapped into the filter importance.
The figure on the right shows how the performance of pre-trained MobileNetV2 changes as its filters are removed based on each filter importance measure.
For each per-layer pruning ratio, the performance degradation happens the least when pre-trained CNNs are pruned with our proposed filter importance measure (BNFI), indicating that an accurate filter importance can be determined by pre-trained BN parameters.
}
\label{fig:1}
\end{figure}
Deep convolutional neural networks (CNNs) have shown remarkable performance in various computer vision tasks~\cite{long2015fully,Ren2015faster,he2016deep}, however CNNs often require large memory storage and expensive computation costs.
Such high demand for memory and computation limits their applications in resource-constrained environment, such as mobile devices.
Among various approaches that tackle this problem, filter pruning approaches have gained attention in recent years as one of prospective approaches for effectively reducing memory storage and computation costs on general tensor processors.

Filter pruning aims to find and remove unimportant filters that do not contribute much to the final output of CNNs.
Owing to the simplicity, filter-weight-based pruning~\cite{li2017pruning,he2019filter} is one of the most widely used approaches in that other model compression methods use it for aiding decision processes on which filters to prune~\cite{yu2018nisp,huang2017condensenet,he2019asymptotic,he2020learning,li2020eagleeye,wang2021convolutional}.
The simplicity is achieved by using readily available information (\eg $\ell_1$-norm of the filter weights) that can be easily computed from off-the-shelf pre-trained networks without processing of training data or pruning-tailored training.
However, relying only on the statistics of filters that have relatively indirect information about CNN outputs, compared to gradients or output feature maps, can render filter-weight-based pruning rather limited~\cite{ye2018rethinking,molchanov2019importance}, resulting in a significant degradation in performance after pruning.

Several works have shifted the attention to other information that is more descriptive with respect to the final output of CNNs, such as activation outputs~\cite{hu2016network,he2017channel, luo2017thinet, lin2020hrank} or gradients of filter weights~\cite{molchanov2019importance, molchanov2017pruning}.
Although using such more informative statistics has led to higher final performance after pruning, they have sacrificed the simplicity in the process.
In particular, they require a number of network feed-forward processes using training data to obtain activation outputs or gradient values.
As such, these data-dependent methods are not suitable as off-the-shelf filter pruning criteria that other advanced pruning algorithms~\cite{yu2018nisp,huang2017condensenet,he2019asymptotic,he2020learning,li2020eagleeye,wang2021convolutional} could depend on.

In this work, we claim that rich information regarding the filter importance can be obtained directly from pre-trained networks without processing of training data or pruning-tailored training.
In particular, we show that the distribution of activation outputs can be estimated with BN parameters of pre-trained networks. 
Such estimation allows us to propose an accurate yet simple-to-compute filter pruning criterion, dubbed as BNFI (\textbf{B}atch \textbf{N}ormalization tells you which \textbf{F}ilter is \textbf{I}mportant), using the pre-trained BN parameters, as illustrated in Figure~\ref{fig:1}.

To validate the effectiveness of the proposed method, we conduct extensive experiments on CIFAR-10 and ImageNet dataset with various network architectures.
We show that the proposed criterion can sort the filters in order of importance more accurately than the filter-weight-based methods and even the data-dependent methods, especially on MobileNetV2~\cite{sandler2018mobilenetv2}.
Combined with a simple greedy per-layer pruning ratio search strategy using the proposed filter importance, the proposed method also achieves better or competitive results after fine-tuning, compared to the existing complex learning-based pruning methods.

We summarize the contributions of this paper as follows:
\begin{itemize}
    \item Without data-dependent computation, the proposed method can estimate the feature distribution of activation outputs using the BN parameters, enabling the measurement of filter importance in terms of the BN parameters.
    \item Experimental results without fine-tuning stages demonstrate that the proposed method outperforms the existing filter pruning methods especially on MobileNetV2.
    \item Using the proposed filter importance, we propose a simple greedy per-layer pruning ratio search strategy and show comparable or better performance after fine-tuning compared to the complex learning-based methods.
\end{itemize}

%% file: main/related_work.tex
\section{Related work}\label{rw}
Filter pruning aims to eliminate unnecessary filters in CNNs.
The filter pruning methods can be roughly categorized into two groups based on whether they perform pruning-tailored training or importance estimation.
The former aims to induce redundant filters via a specialized learning framework while the latter aims to estimate the importance of each filter with respect to the performance of pre-trained networks.
This work can be considered as one of the importance estimation methods, which we further classify them into three groups: filter-weight-based methods, activation-based methods, and gradient-based methods.

The filter-weight-based methods~\cite{li2017pruning,he2019filter} use the filter weight values to estimate the filter importance.
Li~\etal~\cite{li2017pruning} determine the filter importance via $\ell_1$-norm of the filter weights.
He~\etal~\cite{he2019filter} propose to prune the filters near the geometric median of the filters in each layer.
These filter-weight-based methods are widely applicable since the filter importance can be easily evaluated from a pre-trained network without any data-dependent computation.
Owing to its easy-access property, many works adopt the filter-weight-based criterion onto their model compression frameworks~\cite{yu2018nisp,huang2017condensenet,he2019asymptotic,he2020learning,li2020eagleeye,wang2021convolutional}.

The major drawback of the filter-weight-based methods is that they have resorted to simple but indirect measures and neglected other operations in CNNs, such as batch normalization (BN) and non-linear activation function, that can impact the final output of CNNs more directly.
To overcome this problem, the activation-based methods~\cite{hu2016network,luo2017thinet,he2017channel,lin2020hrank,dubey2018coreset} and the gradient-based methods~\cite{dong2017learning,molchanov2019importance} focus on the activation outputs and gradient values of filters.
He~\etal~\cite{he2017channel} and Luo~\etal~\cite{luo2017thinet} find redundant activation channels via complex optimization methods that involve training data.
Dubey~\etal~\cite{dubey2018coreset} prune the filters with the low $\ell_1$-norm of corresponding activation values that are computed over training data before their proposed corset-based compression stage.
Molchanov~\etal~\cite{molchanov2019importance} compute the gradient value of each weight through several network feed-forward processes and estimate the importance of filters via Taylor expansions.
Despite their notable pruning performance, these data-dependent methods suffer from the heavy computation required to find unimportant filters, limiting their applicability.

In this work, we claim that the accurate filter importance can be measured without relying on data for the heavy computation of activation values or optimization processes.
Specifically, we show that the activation outputs (output feature maps of non-linear activation functions) can be estimated using BN parameters of pre-trained networks.

\begin{figure*}[h]
\begin{center}
\includegraphics[width=1\textwidth]{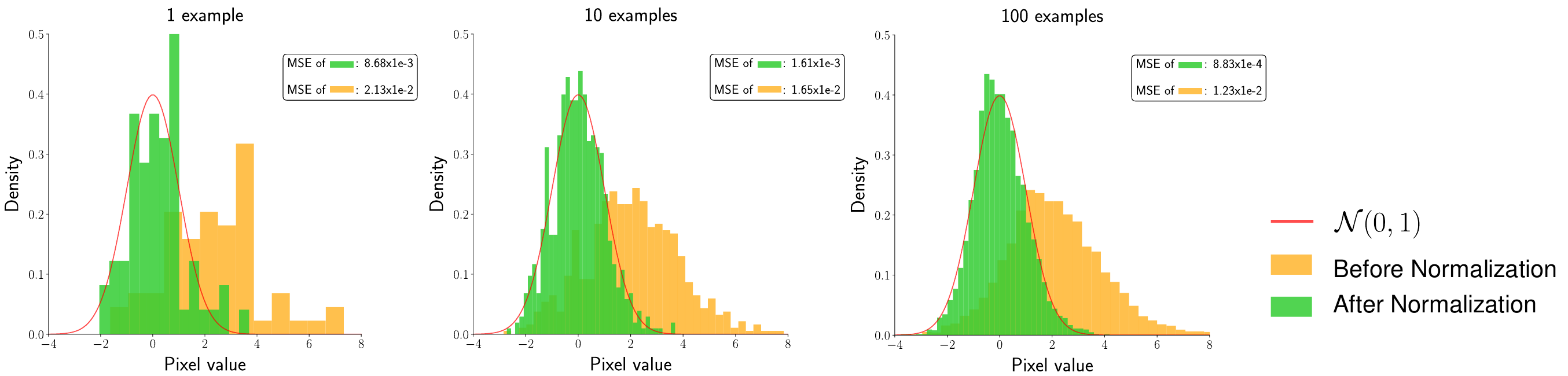}
\end{center}
\vspace{-0.3cm}
\caption{Plots of probability density function of standard normal distribution and normalized histogram of pixel values before normalization and after normalization for a different number of input images. These results are obtained from a channel in a certain layer in ResNet-56 on 1 (left), 10 (middle), and 100 (right) CIFAR-10 test images.
The MSE(Mean Squared Error) score is measured between the histogram and the Gaussian probability distribution function.
The average MSE score after normalization for all channels in the same layer is $4.2 \times 10^{-4}$ on 100 input images.}
\vspace{-0.5cm}
\label{fig:2}
\end{figure*}

Our work may be considered to be similar to sparsity learning methods that use BN layers to guide filter pruning~\cite{liu2017learning,ye2018rethinking,zhao2019variational,kang2020operation}. 
However, they still require training to use BN parameters to intentionally induce redundant activation channels.
Furthermore, these methods do not consider the impact of non-linear activation functions into their proposed filter importance~\cite{liu2017learning,ye2018rethinking,zhao2019variational} or only focus on ReLU activation function~\cite{kang2020operation}.
By contrast, we show that an accurate and flexible filter importance measure can be achieved by using BN parameters of pre-trained networks. 
Specifically, our method is applicable to general activation functions and considers the impact of activation functions for more accurate estimation of the filter importance.

%% file: main/method.tex
\section{Method}\label{method}

Motivated by our observation that the batch normalization (BN) parameters of pre-trained networks contain meaningful information about the activation outputs, we propose to measure the importance of filters in terms of pre-trained BN parameters without relying on data.
In the following subsections, we define the filter pruning problem in Section~\ref{sec:backgroud}, estimate the distribution of activation outputs and define the filter importance using the BN parameters in Section~\ref{definition}, and propose a simple greedy method to find per-layer pruning ratios using the proposed filter importance.

\subsection{Background}

\label{sec:backgroud}
\paragraph{Batch normalization.}
Batch normalization (BN)~\cite{Ioffe2015batch} layer is a commonly used module to facilitate the training process of deep neural networks. 
BN is often placed after the convolution operation whose output feature map is denoted as $\boldsymbol{x} \in \mathbb{R}^{B\times C\times H \times W}$, where $B$, $C$, $H$, and $W$ denote the size of mini-batch, the number of channels, and height and width of the feature map, respectively.
Batch normalization performs normalization and affine transformation for each $i$-th channel $\boldsymbol{x}_i$:
\begin{equation}\label{eq:1}
\hat{\boldsymbol{x}}_i = \frac{\boldsymbol{x}_i - \mu_{B}}{\sqrt{\sigma^2_{B}+\epsilon}},
\end{equation}
\begin{equation}\label{eq:2}
\boldsymbol{x}_{BN,i}=\gamma_i \cdot \hat{\boldsymbol{x}}_i+\beta_i,
\end{equation}
where $\mu_{B}$ and $\sigma_{B}$ respectively denote the mean and standard deviation values of $\boldsymbol{x}_i$ over a mini-batch of size $B$; $\epsilon$ is an arbitrary small constant for numerical stability; and $\gamma_i$ and $\beta_i$ are learnable parameters for each $i$-th channel, respectively.

\vspace{-0.3cm}
\paragraph{Activation function.}
In commonly used modern CNNs, such as VGGNet~\cite{simonyan2015very} and ResNet~\cite{he2016deep}, a BN layer is followed by a non-linear activation layer.
Let $g(\cdot)$ denote the non-linear activation function. 
Then, the final activation output $\boldsymbol{z}$, which has the same dimension as $\boldsymbol{x}$, is formally given by,
\begin{equation}\label{eq:3}
\boldsymbol{z}_{i}=g(\boldsymbol{x}_{BN,i}).
\end{equation}
Note that the activation output $g(\boldsymbol{x}_{BN,i})$ is fed into the next convolutional layer as input.

\vspace{-0.3cm}
\paragraph{Problem definition.}
Let $\boldsymbol{\mathcal{W}}$ denote a set of filters of a convolutional layer.
When pruning $k$ filters in the layer, the optimal set of $k$ unimportant filters is defined as follows:
\begin{equation}\label{eq:4}
    \boldsymbol{\mathcal{W}}_k^s \equiv \underset{\boldsymbol{\mathcal{W}}_k}{\mathrm{argmin}} \: \mathcal{L}(\boldsymbol{D}, \boldsymbol{\mathcal{W}}-\boldsymbol{\mathcal{W}}_k),
\end{equation}
where $\boldsymbol{D}$, $\mathcal{L}$ and $\boldsymbol{\mathcal{W}}_k$ denote the target dataset, the target loss function, and a set of filters with the size of $k$, respectively.
(In Equation~\eqref{eq:4}, the filters of the other layers remain intact.)
Unfortunately, finding $\boldsymbol{\mathcal{W}}_k^s$ requires going through measuring the performance of all possible networks after pruning $k$ filters of the given layer ($\binom{n}{k}$ network inferences), which are intractable computations particularly in recent heavy networks or dataset.
Therefore, many methods have attempted to approximately sort the filters in the order of the estimated relative importance of each filter to obtain the approximate set of $k$ unimportant filters: 
\begin{equation}\label{eq:5}
    \boldsymbol{\mathcal{W}}_k^s \approx \{\mathcal{W}\mid \mathcal{I}_\mathcal{W} \leq \mathcal{I}^k\},
\end{equation}
where $\mathcal{I}_\mathcal{W}$ and $\mathcal{I}^k$ denote the importance of filter $\mathcal{W}$ and $k$-th smallest importance, respectively.
Among them, the activation-based methods focus on the activation outputs to determine the filter importance:
\begin{equation}\label{eq:6}
    \mathcal{I}_{\mathcal{W}_i} \equiv \mathcal{I}(\boldsymbol{z}_i),
\end{equation}
where $\mathcal{I}(\cdot)$ denotes an importance evaluation function.

\subsection{Filter importance from BN parameters}\label{definition}
\paragraph{Gaussian assumption.}
We start with the assumption on the output distribution of a BN layer.
By Equation~\eqref{eq:1}, the input of a BN layer $\boldsymbol{x}_i$ is normalized to $\hat{\boldsymbol{x}}_i$ each element of which, $\hat{x}_i$, has a mean of 0 and a standard deviation of 1.
Under the assumption that the batch size $B$ is sufficiently large, we can approximate the distribution of $\hat{x}_i$ as the Gaussian distribution according to the Central Limit theorem, $\hat{x}_i \sim \mathcal{N}(0,1)$.
Figure~\ref{fig:2} demonstrates that our assumption holds well with a larger batch size.

\vspace{-0.3cm}
\paragraph{Estimating the distribution of activation outputs.}
Under the Gaussian assumption, we can formulate the outputs of subsequent operations in terms of $\gamma_i$ and $\beta_i$.
The following affine transformation in Equation~\eqref{eq:2} shifts and scales the mean and the standard deviation by a factor of $\beta_i$ and $\lvert\gamma_i\rvert$, respectively, leading to $x_{BN,i} \sim \mathcal{N}(\beta_i,\gamma_i^2)$.
Since we can estimate the distribution of $x_{BN,i}$ in terms of $\gamma_i$ and $\beta_i$, we also can estimate that of the activation outputs using BN parameters.
For example, if the activation function is ReLU, $g(x) = max(x,0)$, $z_i$ can be estimated by the truncated Gaussian distribution.
Finally, $x_{BN,i}$ is transformed by a non-linear activation function $g$, producing $z_{i}=g(x_{BN,i})$. 
This suggests that the distribution of activation outputs can be estimated by using the BN parameters since $g$ is known in a target pre-trained network.

\vspace{-0.3cm}
\paragraph{Definition of filter importance.}
With the Gaussian assumption and the subsequent formulation of activation output distributions in terms of BN parameters, we propose a novel definition of filter importance using the estimated activation output distribution.
For our definition of filter importance, we assume that an activation output channel with large absolute values is important because it can have significant impact on the output of the subsequent layers and thus the final output~\cite{luo2017thinet,dubey2018coreset,he2017channel}.
From this assumption, we define the filter importance as \textit{the expectation of absolute values}:
\begin{equation}\label{eq:7}
\mathcal{I}(\boldsymbol{z}_i) \equiv \int_{-\infty}^\infty \lvert g(z) \rvert \cdot f(z;\beta_i,\lvert\gamma_i\rvert)\,dz,
\end{equation}
where $f(\cdot;\beta,\gamma)$ denotes a Gaussian probability density function with a mean of $\beta$ and a standard deviation of $\gamma$.
For computational feasibility, the infinite integration bound can be replaced with small values (e.g. $\pm 5$) with a negligible error and the integration can be easily computed with libraries, such as \textit{SciPy}.

\vspace{-0.3cm}
\paragraph{Filter importance on sparse activation}
Although the proposed filter importance in Equation~\eqref{eq:7} can be applied to any activation function, we empirically found that it is better to consider the sparsity (ratio of zeros) of activation outputs when there is high sparsity on activation outputs (e.g. when the activation function is ReLU).
When the sparsity is high, meaningful information that could be present in non-zero values is dominated by the large number of zero values according to the filter importance defined in Equation~\eqref{eq:7}.
In other words, in the case of spare activation outputs, the corresponding filter should be considered to be important if the expectation of non-zero values is large enough, while it is considered to be unimportant according to Equation~\eqref{eq:7} due to the dominant number of zero values.
Therefore, we slightly modify the definition of filter importance in Equation~\eqref{eq:7} to reduce the impact of the zero values in the case of high sparsity:
\begin{equation}\label{eq:8}
\begin{split}
\mathcal{I}(\boldsymbol{z}_i) &\equiv \int_{-\infty}^\infty \lvert g(z) \rvert \cdot \frac{f(z;\beta_i,\lvert\gamma_i\rvert)}{\mathcal{N}}\,dz,\\
where \:\:\: \mathcal{N} & = \int_{g(z) \neq 0}f(z;\beta_i,\lvert\gamma_i\rvert)\,dz,\\
\end{split}
\end{equation}
Equation~\eqref{eq:8} can be seen as a conditional expectation on positive regions.
Note that Equation~\eqref{eq:8} becomes Equation~\eqref{eq:7} when the activation function does not induce sparsity.
Experiments on the impact of sparsity are provided in the supplementary material.


\begin{figure*}[t]
\begin{center}
\subfloat[Results on CIFAR-10]{
\includegraphics[width=0.31\textwidth]{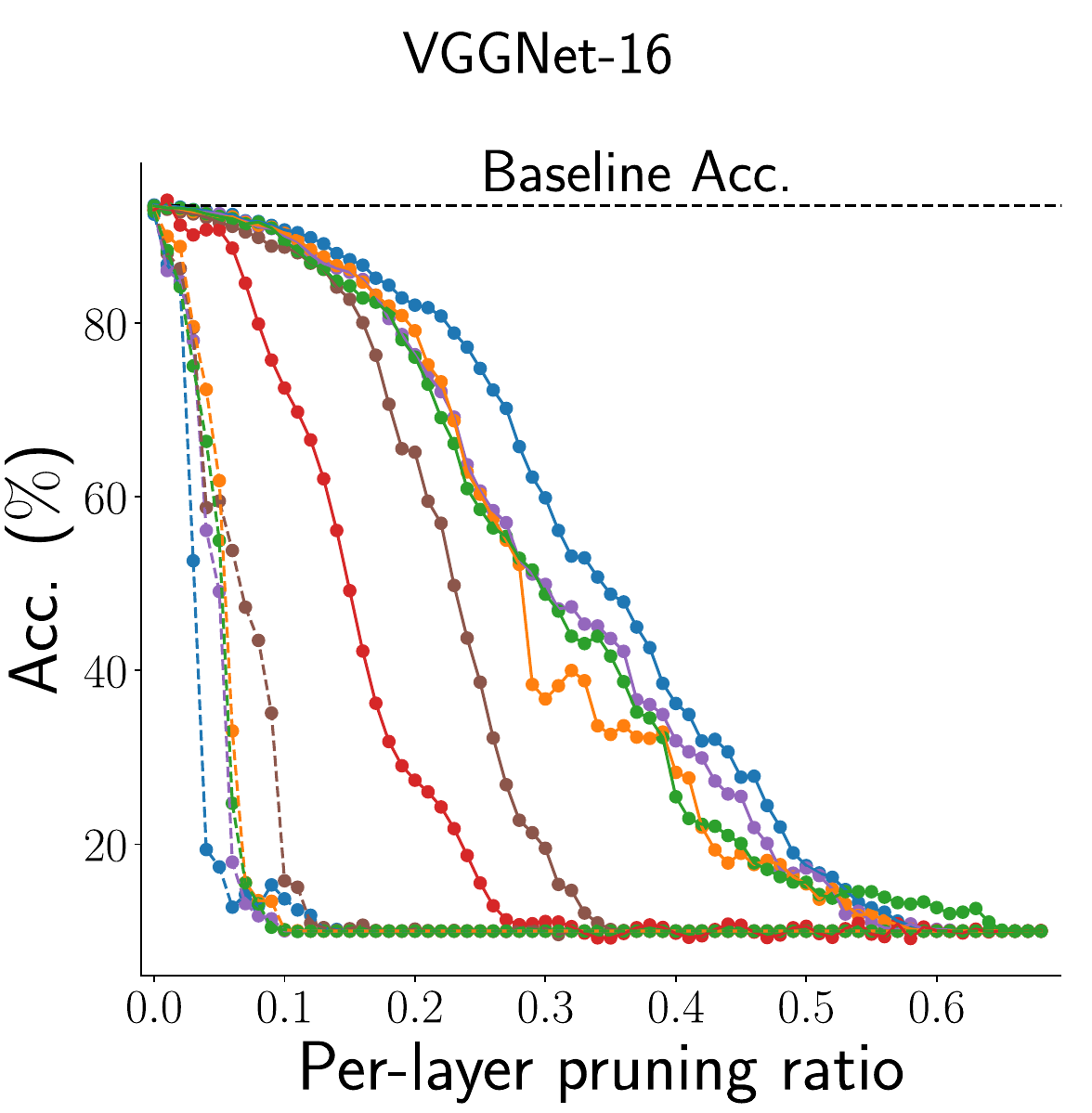}
\label{fig:cifar_woft}}
\subfloat[Results on ImageNet]{
\includegraphics[width=0.62\textwidth]{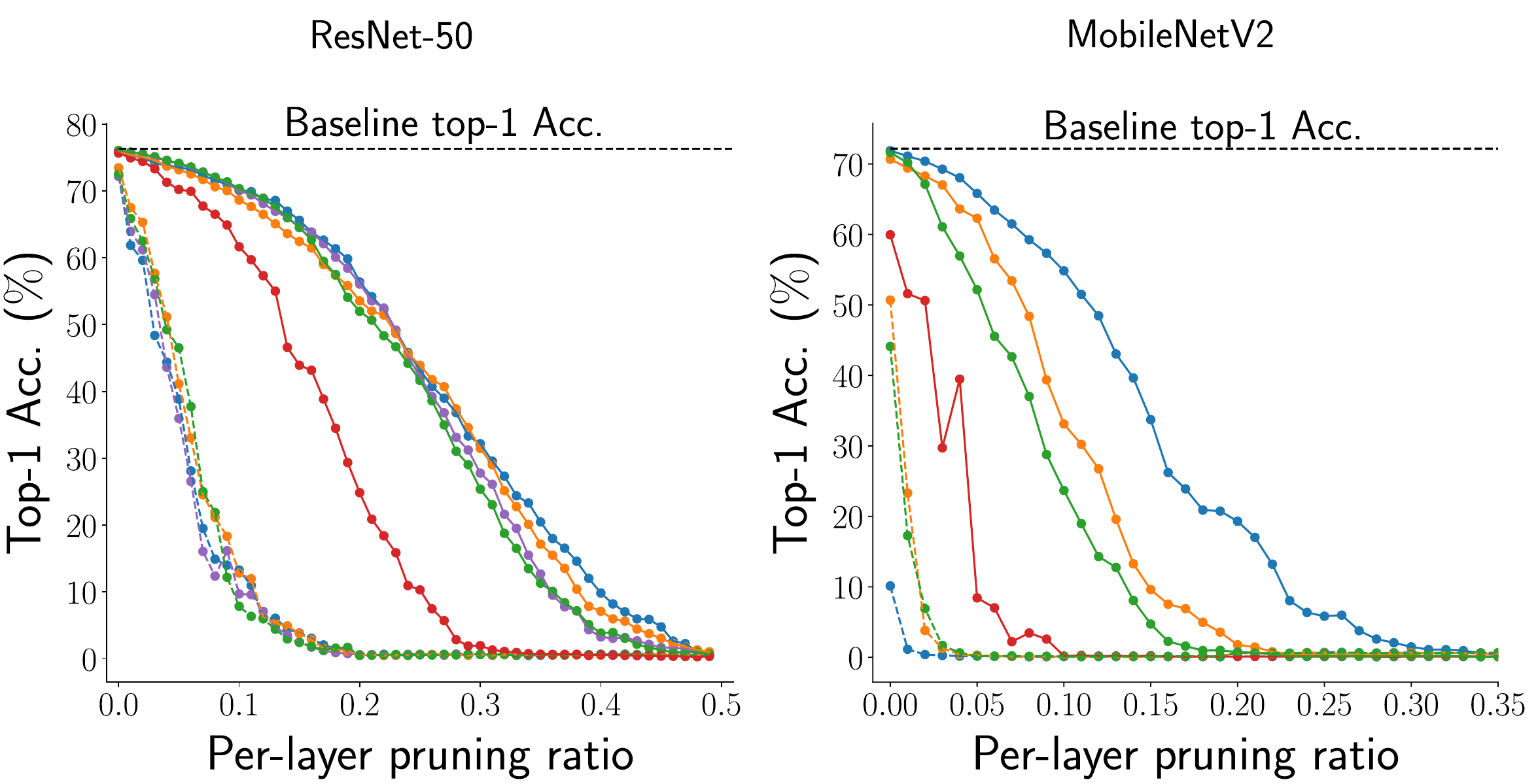}
\label{fig:imagenet_woft}}

\subfloat{
\includegraphics[width=0.85\textwidth]{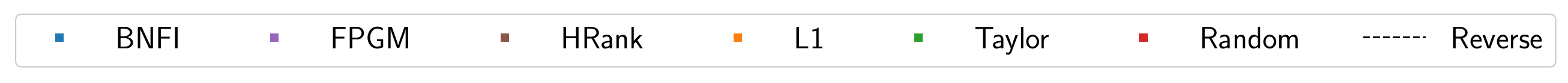}}
\end{center}
\caption{Experimental results on CIFAR-10 and ImageNet without fine-tuning. For each network, we estimate the importance of filters using corresponding method and sort the filters in order of the importance. Then, according to the pruning ratio, we prune the filters in ascending order of importance. For each method, we also present the results on reverse pruning where the filters are removed in descending order of importance.
For the Random method, the filter importance is randomly determined and the results are averaged over 3 experiments.
All layers of a pruned network have the same pruning ratio, which is represented in the x-axis.}
\label{fig:3}
\end{figure*}

\subsection{Per-layer pruning ratio search}\label{pr_search}
\label{ratio}
The performance of a pruned network is largely affected by the structure of the resulting network architecture (e.g. the number of remaining filters in each layer)~\cite{liu2019rethinking,li2020eagleeye,liu2019metapruning}.
Therefore, it is important to carefully design decision processes on how many filters to prune for each layer.
However, filters are interdependent in that removing any filter of a layer can affect other layers.
As it is computationally infeasible to find an optimal number of filters to prune for each layer due to the interdependence and the large number of filters, we propose a greedy method to determine how many filters to prune in each layer based on the proposed filter importance measure.

In particular, we determine the number of filters to prune for each layer based on the contribution of each layer to the final performance of a pre-trained network.
The contribution of the $i$-th layer is measured as the performance degradation $\delta_i$ after pruning a certain portion (pruning ratio $r_i$) of filters in the layer (where filters with the least importance are pruned first according to Equation~\eqref{eq:8}), where $\delta_i$ can be calculated as the difference between the validation accuracy of a pre-trained network $V_{base}$ and that of a resulting pruned network $V'_i$:
\begin{equation}\label{eq:9}
\delta_i =  |V_{base} - V'_i(r_i)|.
\end{equation}
Based on the relation between the performance degradation and pruning ratio outlined in Equation~\eqref{eq:9}, we find a pruning ratio $r_i$ for each $i$-th layer that gives a pruned network whose performance degradation closely matches $\delta_i$, which is given as a hyper-parameter.
To find $r_i$ efficiently, we conduct an iterative algorithm similar to binary search.
Given a lower bound of $l$ and an upper bound of $u$, we measure the validation accuracy $V'_i(\frac{l+u}{2})$, compare the degraded performance to $\delta_i$, and modify $l$ or $u$ according to the comparison result.
After 5 iteration, we select $\frac{l+u}{2}$ as a resulting pruning ratio.

%% file: main/experiment.tex
\section{Experiment}\label{experiment}

\subsection{Experimental details}
\paragraph{Datasets.}
We use CIFAR-10~\cite{cifar10} and ImageNet datasets~\cite{deng2009imagenet} to evaluate the classification performance of the baseline and pruned models.
CIFAR-10 are widely used dataset that contains 50,000 training and 10,000 test images with 10/100 classes and $32 \times 32$ resolution.
ImageNet is a large-scale and challenging dataset which consists of 1,281,167 training and 50,000 validation images with 1,000 classes.
For data augmentation, we conduct random cropping and flipping on all training sets and center cropping for the validation set of ImageNet.

\paragraph{Models.}
We use the recent commonly used CNNs models, such as VGGNet~\cite{simonyan2015very}, ResNet \cite{he2016deep,he2016identity}, and MobileNetV2~\cite{sandler2018mobilenetv2}.
For the experiments on CIFAR-10 dataset, we use the light version of VGGNet-16, as in \cite{li2017pruning} and~\cite{liu2017learning}.
For the experiments on ImageNet dataset, we use ResNet-50 and MobileNetV2 to validate the effectiveness of the proposed method on such complex or light models.

\paragraph{Training Settings.}
For VGGNet-16 and ResNet-50, we train the baseline models with the batch size of 64 for 160 epochs on CIFAR-10 and the batch size of 256 for 120 epochs on ImageNet datasets.
In the fine-tuning stage, the settings are the same except that fewer epochs are used (120 for CIFAR-10 and 80 for ImageNet datasets).
We use the stochastic gradient descent algorithm for optimizer with weight decay $10^{-4}$ and momentum 0.9, and the learning rate is initially set to 0.1 and decayed by 0.1 at one-half and one-quarter of total epochs.
In the case of MobileNetV2, we use weight decay $5\times10^{-4}$ and momentum 0.9, and the learning rate is scheduled by cosine annealing scheduler, updated from 0.05 to 0.

\paragraph{Evaluation Method.} For the evaluation of pruning performance, we use three metrics: the accuracy drop with respect to the accuracy of baseline models, the ratio of the pruned model parameters, and computational complexity. 
Instead of the final accuracy of the pruned model, we use the accuracy drop as a performance measure of the pruned model since the results of previous methods have slightly different baseline accuracy.
We use floating point operations (FLOPs) as the measurement of computational complexity. 
As for the ratio of the pruned parameters, we count the number of parameters and FLOPs only for the convolutional, following the settings from~\cite{he2019filter,he2018lsoft}
Using the three metrics, we mainly compare the proposed pruning method to the importance estimation method~\cite{li2017pruning,he2019filter,lin2020hrank,molchanov2019importance} without fine-tuning settings to demonstrate that the proposed filter importance BNFI better identifies the network redundancy.

\paragraph{Implementation details.}
For both ResNet-50 and MobileNetV2, where each residual block has 1x1-3x3-1x1 convolutional layers, we only prune the 3x3 convolutional layer since the computations of 1x1 convolutional layers are relatively light.
To reproduce the results of the gradient-based method~\cite{molchanov2019importance}, we used the entire training data to accumulate the gradient values.
We used the available source code to reproduce the activation-based method~\cite{lin2020hrank}.
When applying the proposed per-layer pruning ratio strategy, we use the training accuracy instead of the validation accuracy since validation data is not available.
In the case of CIFAR-10 dataset, we obtain the final pruned network through several searching-pruning-fine-tuning processes to carefully prune the network.
We use the entire training data for CIFAR-10 and 50,000 selected training data for ImageNet for the strategy.
All experiments are conducted on RTX 2080Ti GPUs.

\begin{figure}[t]
\begin{center}
\includegraphics[width=0.95\linewidth]{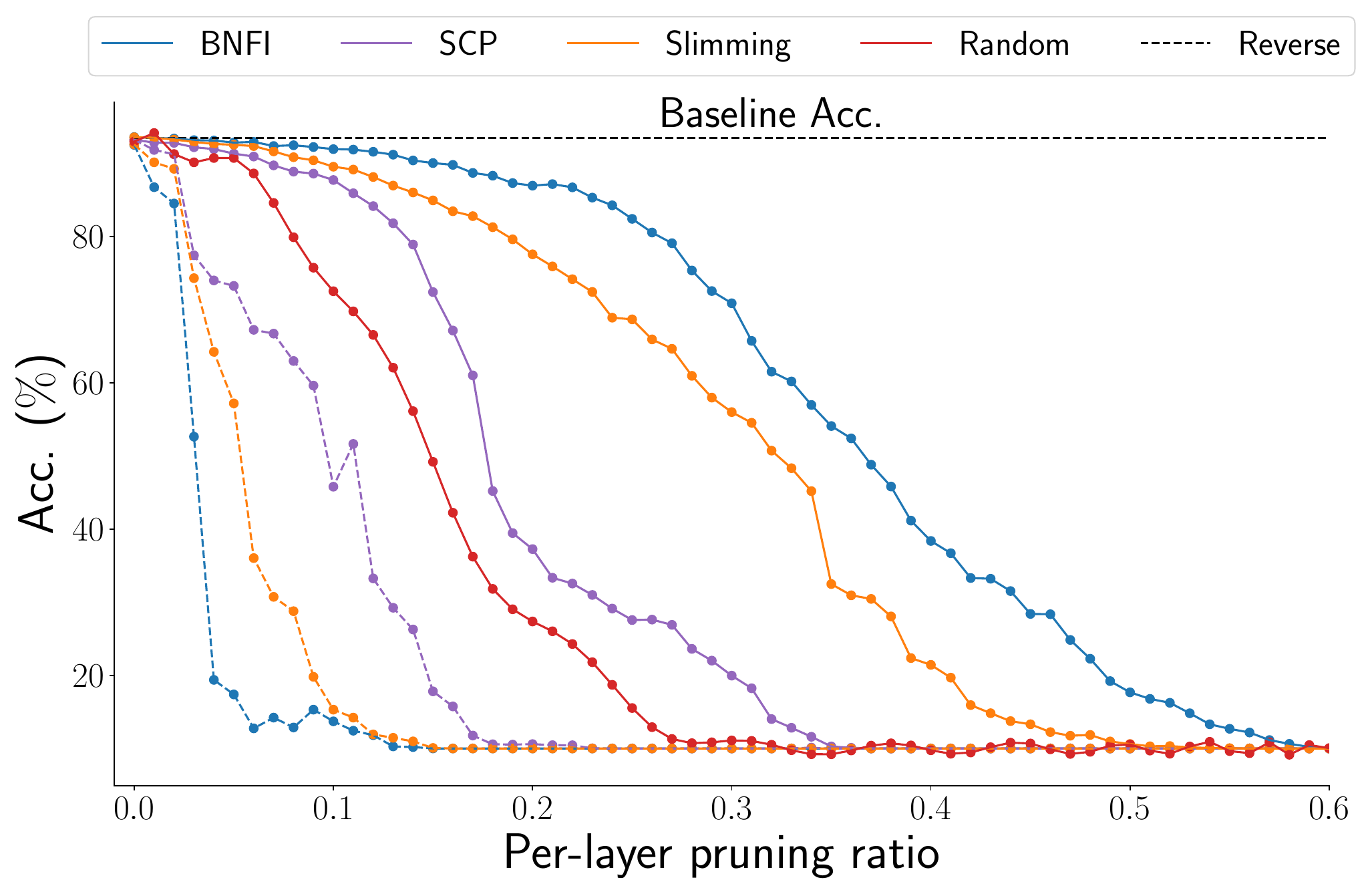}
\end{center}
\begin{center}
\end{center}
\vspace{-1.1cm}
\caption{Comparison with the BN-based methods without fine-tuning. All details are same with the experiments on Figure~\ref{fig:3}. The baseline network and the dataset is VGGNet-16 and CIFAR-10.
}
\label{fig:4}
\end{figure}

\subsection{Pruning without fine-tuning}
In this subsection, we present the pruning results without fine-tuning stage to demonstrate the effectiveness of the proposed filter importance for determining redundant filters in pre-trained networks.
We compare BNFI with the existing filter-weight-based methods (L1 and FPGM)~\cite{li2017pruning,he2019filter}, the activation-based method (HRank)~\cite{lin2020hrank}, and the gradient-based method (Taylor)~\cite{molchanov2019importance}.
Note that HRank and Taylor are data-dependent methods.
For simplicity, let DOI and AOI denote ``pruning in a descending order of importance" and ``pruning in an ascending order of importance", respectively.
\paragraph{Results on CIFAR-10.}
Figure~\ref{fig:cifar_woft} shows the results on CIFAR-10.
Across most of the pruning ratios, the performance degradation of BNFI is less in AOI and more in DOI by a significant margin.
These results mean that BNFI can find both important filters and unimportant filters better than the other methods.
We want to note that even compared to HRank and Taylor, which are data-dependent methods, BNFI achieves outstanding results.
These results demonstrate that the proposed method can find the unimportant filters more accurately even without data-driven information.

\begin{figure}[t]
\begin{center}
\subfloat[Swish]{
\includegraphics[width=0.25\textwidth]{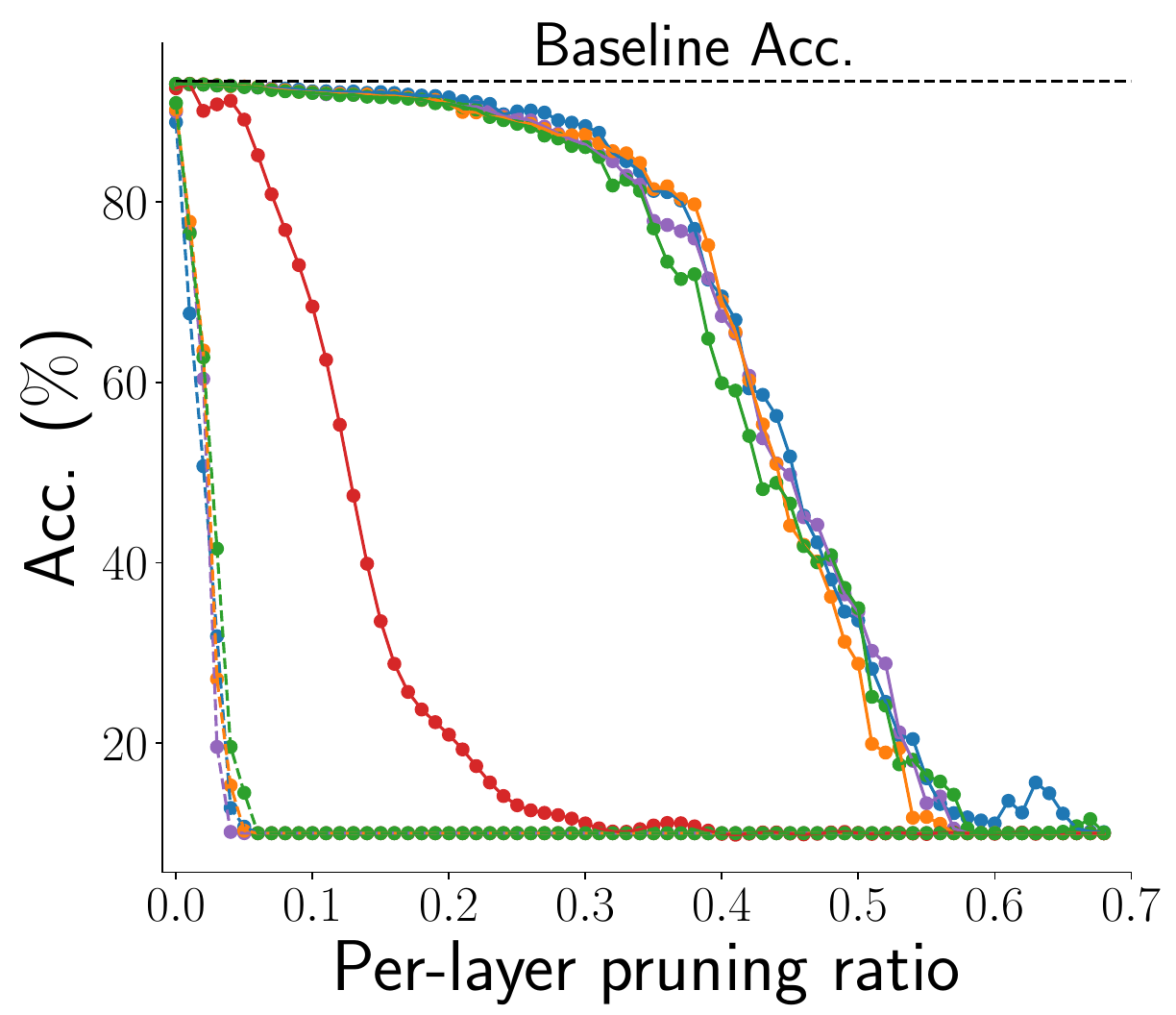}
\label{fig:feature_distribution}}
\subfloat[Leaky ReLU]{
\includegraphics[width=0.25\textwidth]{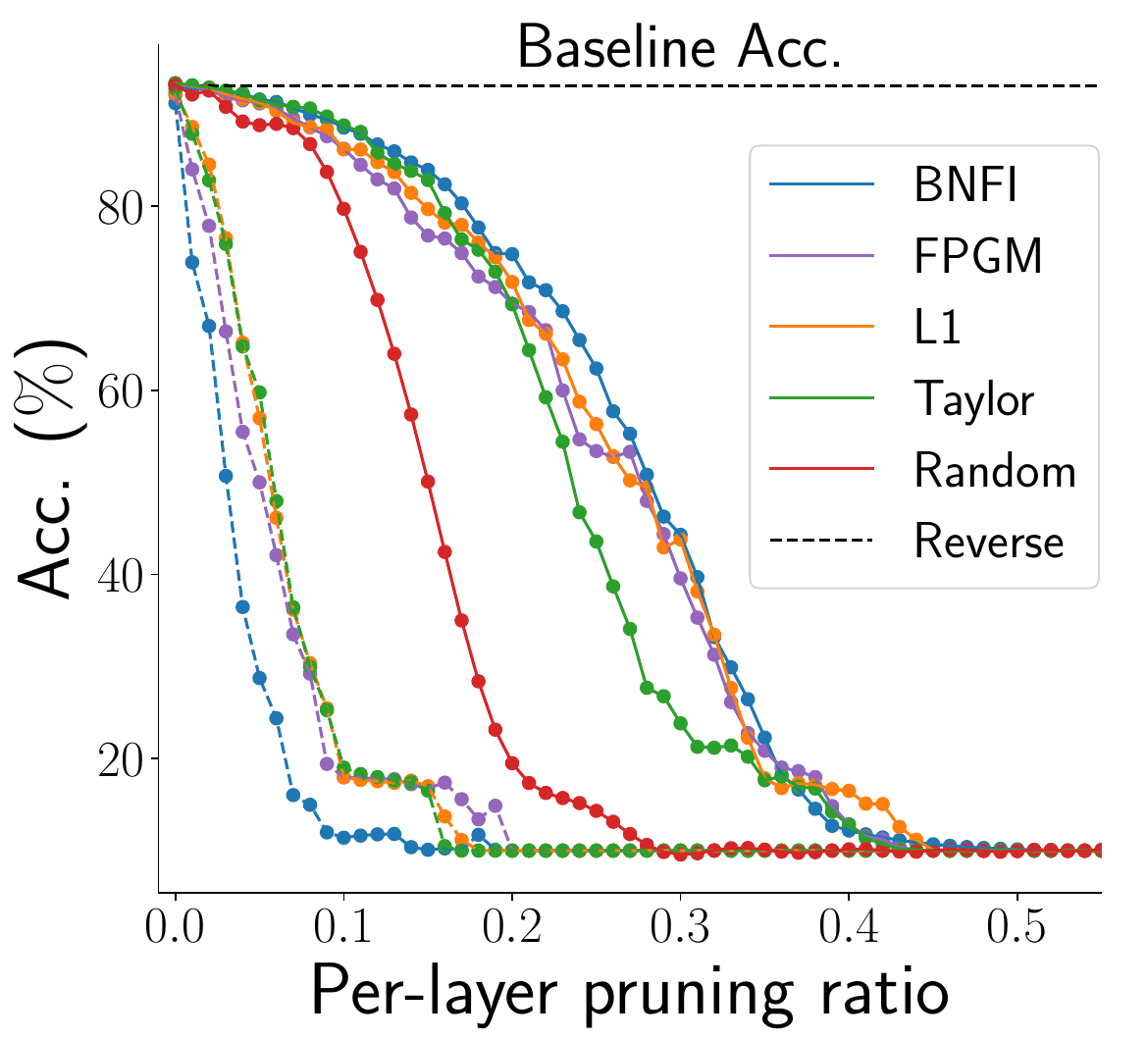}
\label{fig:importance_mapping_by_n}}
\end{center}
\caption{Experimental results on other activation functions. All details are same with the experiments on Figure~\ref{fig:3}. The baseline networks are VGGNet-16 with a different activation function.
}
\label{fig:5}
\end{figure}

\paragraph{Results on ImageNet.}
We conduct experiments on much more complex dataset, ImageNet dataset.
The results are presented in Figure~\ref{fig:imagenet_woft}.
For ResNet-50, BNFI shows almost same or slightly less accuracy drop until the pruning ratio of 0.3 but outperforms the existing methods in pruning ratio larger than 0.3 in AOI, which means the proposed method also works well on such a complex model.
Remarkably, BNFI significantly outperforms L1 and Taylor in both AOI and DOI in MobileNetV2.
In AOI, the accuracy of BNFI in pruning ratio from 0.1 to 0.15 is higher than 30$\%$ points compared to the results of the other methods.
Further, when the pruning ratio is merely 0.01, the accuracy is destructively degraded by BNFI in DOI, 72.19$\%$ to almost 10$\%$ and almost 0$\%$ after pruning ratio of 0.02.
These outstanding results demonstrate that BNFI can accurately identify important and unimportant filters in depth-wise convolutional layers much better than L1 and Taylor.
These experimental results suggest that BNFI can be used to investigate the redundancy in such a recent efficient architecture.

\paragraph{Comparision with the BN-based methods.}
Similar to our work, the existing BN-based methods~\cite{liu2017learning,ye2018rethinking,zhao2019variational,kang2020operation} also determine the importance of filter via BN parameters.
However, our proposed filter importance measure BNFI is a more accurate criterion than other existing BN-based methods, which require an ad-hoc learning process to forcibly induce unimportant filters.
To support our claim, we compare with other BN-based methods in terms of the accuracy without fine-tuning, as illustrated in Figure~\ref{fig:4}.
In AOI, BNFI causes a minor accuracy drop, less than 10$\%$, until the pruning ratio of 0.25.
On the other hand, Slimming and SCP significantly degrade the network performance as the pruning ratio increases, particularly in the case of SCP.
Notably, in the pruning ratio of 0.2, BNFI still maintains the original performance, about 90$\%$, but Slimming and SCP achieve 80$\%$ and 40$\%$, respectively.
In DOI, BNFI causes a destructive damage on the performance, about 20$\%$ in the pruning ratio of 0.05, but Slimming and SCP still show a good accuracy, about 75$\%$.
The overall results demonstrate that BNFI is more accurate than Slimming and SCP, meaning that the proposed method utilizes the concealed information much better than the existing BN-based methods.

\input{Table/table1}

\paragraph{Generalization to other activation functions.}
In this section, we show the proposed filter importance is also effective for other activation functions, such as Swish~\cite{swish} and Leaky ReLU~\cite{leaky}.
To validate the generality, we conduct experiments without fine-tuning on pre-trained networks with different activation functions, which is shown in Figure~\ref{fig:5}.
In the case of Swish, BNFI shows comparable or slightly good performance on average in AOI and almost no accuracy drop until the pruning ratio of 0.2.
These results demonstrate that BNFI can estimate the distribution of activation outputs, and thus the importance quite accurately even when the activation function is Swish.
BNFI shows quite better results on Leaky ReLU.
BNFI causes more steep drop of accuracy than the other methods in all of the pruning ratios in DOI.
In AOI, BNFI shows rather good results in the pruning ratio from 0.2 to 0.3, more than 5$\%$ points than the other methods, demonstrating BNFI also can be applied to Leaky ReLU.
All of these results show that the proposed method is applicable to other activation functions.

\subsection{Pruning with fine-tuning}
Although the focus of this paper is to develop a novel filter importance measure, we compare our method to more complex methods with pruning-tailored learning process to show the potential of BNFI.
We sort the filters of a pre-trained network according to BNFI and search per-layer pruning ratios using the method in Section~\ref{ratio}.
Then, we prune the full network and fine-tune for several epochs.

\vspace{-0.3cm}
\paragraph{Impact on after fine-tuning results.}
Before the discussion on the main results, we want to discuss the correlation between the performance without and with fine-tuning to demonstrate that the final performance is highly affected by which filters to prune, demonstrating the value of our work.
The experimental results are shown in Table~\ref{tab:1}.
The only difference of each result with same pruning ratio is which filters are removed.
As shown in Figure~\ref{fig:3}, the performance without fine-tuning is much better when the filter importance is measured by BNFI.
Similar to the results without fine-tuning, the results after fine-tuning significantly differ in the final accuracy, about 1$\%$ points between the results of BNFI and Reverse.
These results indicate that the accurate filter importance is valuable for achieving higher performance even after fine-tuning.

\input{Table/table2}
\input{Table/table3}
\vspace{-0.3cm}
\paragraph{Results on CIFAR-10.}
Table~\ref{tab:2} shows the experimental results after fine-tuning on CIFAR-10 dataset.
Through several searching-pruning-fine-tuning processes, the proposed method found a very efficient architecture whose number of parameters and FLOPs are pruned by 97.07$\%$ and 74.81$\%$, respectively, which is most efficient result among the presented results.
Despite such a huge reduction ratio, the accuracy drop is merely 0.04$\%$.
These results are outstanding compared to the sparsity learning methods (Slimming, Variational Pruning and SCP) and to the importance estimation methods (L1 and HRank), demonstrating the effectiveness of the proposed filter importance for determining where to prune and how many to prune.

\vspace{-0.3cm}
\paragraph{Results on ImageNet.}
To further evaluate BNFI on a larger dataset, we also conduct experiments on ImageNet, which is presented in Table~\ref{tab:3}.
For ResNet-50, we provide three results with different FLOPs.
BNFI shows an accuracy drop of 0.16 when FLOPs are 3.02G, which is better than the sub-network search method, MetaPruning.
In the intermediate pruning ratio, BNFI acheives an accuracy drop of 0.86, which is outstanding results compared to the other methods.
For the most efficient results, BNFI also shows considerably great results than the complex learning-based methods, such as DSA and Hinge.
The above results demonstrate that the proposed filter importance is effective in determining where to prune and how many to prune, even on such a large dataset and a complex network.
We also present the results on MobileNetV2.
In MobileNetV2, BNFI achieves comparable or better results to the learning-based methods, an accuracy drop of 1.22 with FLOPs of 220M and an accuracy drop of 3.47 with FLOPs of 158M, which is slightly better than the result of DMC.
Considering that the results of the learning-based methods are due to the learning process to learn where and how to prune simultaneously, the results show the potential of BNFI since the two major factors are only determined by BNFI.

%% file: Table/table1.tex
\begin{table}[t]
   \caption{Experimental results on the impact of pruning criterion onto the final performance after fine-tuning. The pre-trained MobileNetV2 is pruned by each criterion and fine-tuned on ImageNet dataset. The pruning ratio is shared across the entire layers. For the Random criterion, we report the mean and standard deviation of 3 experiments. The Reverse method determines the filter importance using the proposed method and prune the filters in order of importance.}
   \centering
   \scalebox{0.85}{
   \begin{tabular}{ccc}
   \toprule[\heavyrulewidth]
          Pruning ratio & Criterion & Acc.($\%$)\\
   \midrule
   \multirow{4}{*}{0.25} & BNFI & $\textbf{71.59}$ \\
   & L1 & $71.50$\\
   & Random & $71.37 (\pm 0.05)$\\
   & Reverse & $70.64$\\
   \midrule
   \multirow{4}{*}{0.5} & BNFI & $\textbf{69.65}$ \\
   & L1 & $69.53$\\
   & Random & $69.35 (\pm 0.06)$\\
   & Reverse & $68.30$ \\
   \bottomrule[\heavyrulewidth]
   \end{tabular}}
   \label{tab:1}
\end{table}

%% file: Table/table2.tex
\begin{table*}[t]
   \caption{Pruning and fine-tuning results on CIFAR-10. 
   The notation "$\downarrow$" means the reduction rate of the corresponding metrics.} 
   \centering
   \scalebox{0.8}{
   \begin{threeparttable}
   \begin{tabular}{llccccc}
   \toprule[\heavyrulewidth]
          Model & Method & Baseline Acc. (\%) & Acc. (\%) & Acc. Drop (\%)& Parameters $\downarrow$ (\%) & FLOPs $\downarrow$ (\%)\\
   \bottomrule[\heavyrulewidth] 
   \multirow{6}{*}{VGGNet-16}&Slimming~\cite{liu2017learning}* & $93.85$ & $92.91$ & $0.94$ & $87.97$ & $48.12$\\
   &Variational Pruning~\cite{zhao2019variational} & $93.25$ & $93.18$ & $0.07$ & $73.34$ & $39.10$\\
   &SCP~\cite{kang2020operation} & $93.85$ & $93.79$ & $0.06$ & $93.05$ & $66.23$\\
   &L1~\cite{li2017pruning} & $93.25$ & $93.40$ & $-0.15$ & $64.00$ & $34.20$\\
   &HRank~\cite{lin2020hrank} & $93.96$ & $93.43$ & $0.53$ & $82.90$ & $53.50$\\
   &\cellcolor{orange!25}\textbf{BNFI (Ours)} & \cellcolor{orange!25}$93.50$ & \cellcolor{orange!25}$93.46$ & \cellcolor{orange!25}$0.04$ & \cellcolor{orange!25}$94.07$ & \cellcolor{orange!25}$74.81$\\
   \bottomrule[\heavyrulewidth]
   \end{tabular}
   \begin{tablenotes}
   \item[*]Reproduced results in~\cite{kang2020operation}.
   \end{tablenotes}
   \end{threeparttable}
   }
   \label{tab:2}
\end{table*}

%% file: Table/table3.tex
\begin{table*}[t]
   \caption{Pruning and fine-tuning results on ImageNet.
   } 
   \centering
   \scalebox{0.8}{
   \begin{tabular}{llcccc}
   \toprule [\heavyrulewidth]
    Model & Method & Baseline Top-1 Acc. (\%) & Top-1 Acc. (\%) & Top-1 Acc. Drop (\%) & FLOPs\\
   \midrule
   \multirow{15}{*}{ResNet-50} & MetaPruning~\cite{liu2019metapruning} & $76.60$ & $76.20$ & $0.40$  & $3.00$G\\
   &\cellcolor{orange!25}\textbf{BNFI (Ours)} & \cellcolor{orange!25}$76.33$ & \cellcolor{orange!25}$76.17$ & \cellcolor{orange!25}$0.16$ & \cellcolor{orange!25}$3.02$G\\
   \cline{2-6}
   &SFP~\cite{he2018lsoft} & $76.15$ & $74.61$ & $1.54$  & $2.39$G\\
   &ThinNet~\cite{luo2017thinet} & $72.88$ & $72.04$ & $0.84$  & $2.90$G\\
   &HRank~\cite{lin2020hrank} & $76.15$ & $74.98$ & $1.17$  & $2.30$G\\
   &Taylor~\cite{molchanov2019importance} & $76.18$ & $74.50$ & $1.68$  & $2.25$G\\
   &DSA~\cite{ning2020dsa} & $76.02$ & $74.10$ & $0.92$ & $2.47$G\\
   &\cellcolor{orange!25} \textbf{BNFI (Ours)} & \cellcolor{orange!25}$76.33$ & \cellcolor{orange!25}$75.47$ & \cellcolor{orange!25}$0.86$ & \cellcolor{orange!25}$2.34$G\\
   \cline{2-6}
   &MetaPruning~\cite{liu2019metapruning} & $76.60$ & $75.40$ & $1.20$  & $2.01$G\\
   &HRank~\cite{lin2020hrank} & $76.15$ & $71.98$ & $4.17$  & $1.55$G\\
   &FPGM~\cite{he2019filter}& $76.15$ & $74.83$ & $1.32$  & $1.92$G\\
   &DCP~\cite{zhuang2018discirmination} & $76.01$ & $74.95$ & $1.06$ & $1.83$G\\
   &Hinge~\cite{li2020group} & $76.10$ & $74.70$ & $1.40$ & $1.88$G\\
   &DSA~\cite{ning2020dsa} & $76.02$ & $74.69$ & $1.33$ & $2.06$G\\
   &\cellcolor{orange!25}\textbf{BNFI (Ours)} & \cellcolor{orange!25}$76.33$ & \cellcolor{orange!25}$75.02$ & \cellcolor{orange!25}$1.29$ & \cellcolor{orange!25}$1.94$G\\
   \midrule
   \multirow{7}{*}{MobileNetV2} & MetaPruning~\cite{liu2019metapruning}& $72.00$ & $71.20$ & $0.80$  & $217$M\\
   &AMC~\cite{he2018amc} & $71.80$ & $70.80$ & $1.00$ & $220$M\\
   &LeGR~\cite{chin2020legr} & $-$ & $71.40$ & $-$ & $224$M \\
   &\cellcolor{orange!25}\textbf{BNFI (Ours)} & \cellcolor{orange!25}$72.19$ & \cellcolor{orange!25}$70.97$ & \cellcolor{orange!25}$1.22$ & \cellcolor{orange!25}$220$M\\
   \cline{2-6}
   &DMC~\cite{gao2020discrete} & $71.88$ & $68.37$ & $3.51$ & $162$M \\
   &LeGR~\cite{chin2020legr} & $-$ & $69.40$ & $-$ & $160$M \\
   &\cellcolor{orange!25} \textbf{BNFI (Ours)} & \cellcolor{orange!25}$72.19$ & \cellcolor{orange!25}$68.72$ & \cellcolor{orange!25}$3.47$ & \cellcolor{orange!25}$158$M\\
   \cline{2-6}
   \bottomrule [\heavyrulewidth]
   \end{tabular}}
   \label{tab:3}
\end{table*}

%% file: main/conclusion.tex
\section{Conclusion}\label{conclusion}
In this paper, we propose a novel filter pruning criterion, BNFI, by extracting the concealed information in the BN parameters.
From the Gaussian assumption, we estimate the distribution of activation outputs, leading to our definition of the filter importance in terms of the BN parameters.
Since BNFI considers the impact of the BN layer and the activation function, it enables more accurate filter pruning decision than the existing filter-weight-based methods, while maintaining the easy-access property.
The experimental results without fine-tuning on various models and datasets demonstrate that BNFI can find unimportant and important filters more accurately than the existing methods, especially on MobileNetV2.
Furthermore, we show that the proposed filter importance can be used to search the per-layer pruning ratio, exhibiting the comparable or better results after fine-tuning compared to the complex learning-based methods.
\vspace{-0.3cm}